\documentclass[runningheads]{llncs}
\usepackage{graphicx}
\usepackage{amsmath,amssymb} 
\usepackage{color}
\usepackage[width=122mm,left=12mm,paperwidth=146mm,height=193mm,top=12mm,paperheight=217mm]{geometry}
\usepackage{multirow}
\usepackage{subcaption}
\begin{document}
\pagestyle{headings}
\mainmatter

\title{Semantic Clustering for Robust Fine-Grained Scene Recognition} 

\titlerunning{Semantic Clustering for Robust Fine-Grained Scene Recognition}

\authorrunning{George et al.}

\author{Marian George\inst{1} \and Mandar Dixit\inst{2} \and G\'{a}bor Zogg\inst{1} \and Nuno Vasconcelos\inst{2}}



\institute{Department of Computer Science,
        ETH Zurich, Switzerland\\
        \email{ mageorge@inf.ethz.ch, gzogg@student.ethz.ch} \and
        Statistical and Visual Computing Lab,
        UCSD, CA, United States\\
        \email{\{mdixit, nvasconcelos\}@ucsd.edu}
}

\maketitle

\begin{abstract}
In domain generalization, the knowledge learnt from one or multiple source domains is transferred to an unseen target domain.
In this work, we propose a novel domain generalization approach for fine-grained scene recognition.
We first propose a semantic scene descriptor that jointly captures the subtle differences between fine-grained scenes, while being robust to varying object configurations across domains. We model the occurrence patterns of objects in scenes, capturing the informativeness and discriminability of each object for each scene. We then transform such occurrences into scene probabilities for each scene image. Second, we argue that scene images belong to hidden semantic topics that can be discovered by clustering our semantic descriptors. 
To evaluate the proposed method, we propose a new fine-grained scene dataset in cross-domain settings. Extensive experiments on the proposed dataset and three benchmark scene datasets show the effectiveness of the proposed approach for fine-grained scene transfer, where we outperform state-of-the-art scene recognition and domain generalization methods.

\end{abstract}

\section{Introduction}
Scene classification is an important problem for computer vision.  Discovering the discriminative aspects of a scene in terms of its global representation, constituent objects and parts, or their spatial layout remains a challenging endeavor. Indoor scenes \cite{indoor_scenes} are particularly important for applications such as robotics.  They are also  particularly challenging, due to the need to understand images at multiple levels of the continuum between things and stuff~\cite{stuff}. Some scenes, such as a garage or corridor, have a distinctive holistic layout.
Others, such as a bathroom, contain unique objects. All of these challenges are aggravated in the context of fine-grained indoor scene classification. Fine-grained recognition targets the problem of sub-ordinate categorization. While it has been studied in the realm of objects, e.g. classes of birds \cite{ucsd_birds}, or flowers \cite{oxford_flowers}, it has not been studied for scenes.

In real-world applications, vision systems are frequently faced with the need to process  images  taken under very different imaging conditions than those in their training sets. This is frequently called the cross-domain setting, since the domain of test images is different from that of training. For example, store images taken with a smartphone can differ significantly from those found on the web, where most image datasets are collected. The variation can be in terms of the objects displayed (e.g. the latest clothing collection), their poses, the lighting conditions, camera characteristics, or proximity between camera and scene items. It is well known that the performance of vision models can degrade significantly due to these variations, which is known as the dataset bias problem \cite{dataset_bias,bias_perronnin_2010}. 

To address the dataset bias problem, many domain adaptation  \cite{domain_adaptation_survey} approaches have been proposed \cite{domain_adaptation_1,domain_adaptation_2,domain_adaptation_3,domain_adaptation_4} to reduce the mistmatch between the data distributions of the training samples, referred to as source domain, and the test samples, referred to as the target domain. In domain adaptation, target domain data is available during the training process, and the adaptation process needs to be repeated for every new target domain. A related problem is \textit{domain generalization}, in which the target domain data is unavailable during training \cite{dica,undo_bias,lre_svm,mtae,wsdg}.
Such problem is important in real-world applications where different target domains may correspond to images of different users with different cameras.

\begin{figure*}[t]
\begin{center}
\includegraphics[width=\linewidth]{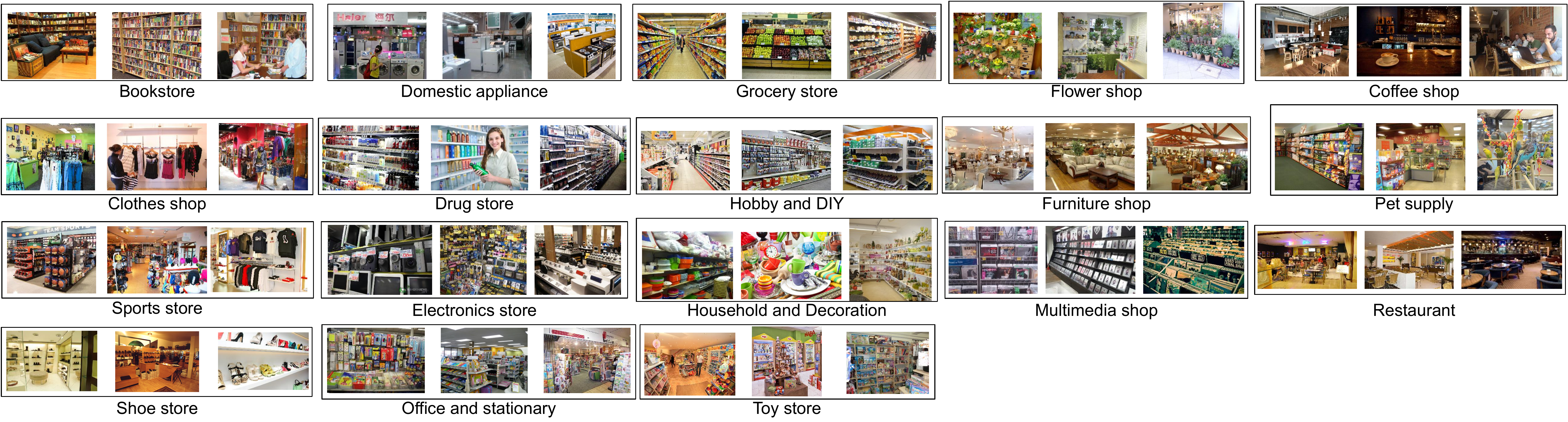}
\end{center}
   \caption{An overview of the proposed fine-grained scene \textit{SnapStore} dataset.
   The dataset contains $18$ store categories that are closely related to each other.
	For each category, $3$ training images are shown. Some categories are significantly visually similar with very confusing spatial layout and objects. Other store classes have widely varying visual features, which is difficult to model.
	}
\label{fig:dataset}
\end{figure*}

In this work, we study the problem of domain generalization for fine-grained scene recognition by considering store scenes. As shown in Figure  \ref{fig:dataset}, store classification frequently requires the discrimination between classes of very similar visual appearance, such as a drug store vs. a grocery store. Yet, there are also classes of widely varying appearance, such as clothing stores. This makes the store domain suitable to test the robustness of models for scene classification.

\begin{figure}[t]
 \centering
   \includegraphics[width=0.95\linewidth]{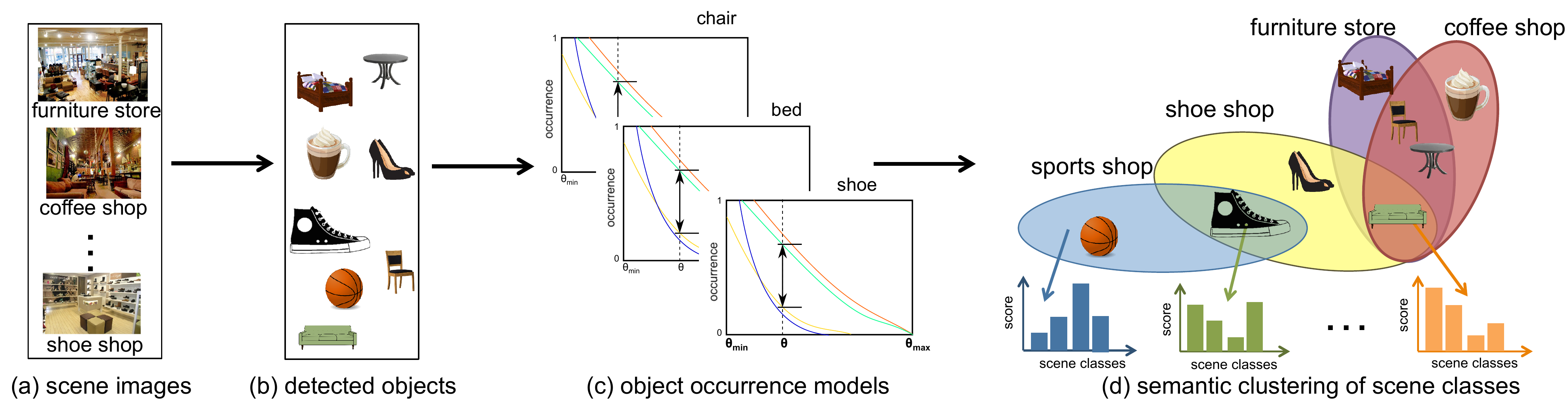}
   \caption{Overview of our semantic clustering approach. (a) scene images from all scene classes are first projected into (b) a common space, namely object space. (c) Object occurrence models are computed to describe conditional scene probabilities given each object. The maximal vertical distance between two neighboring
     curves at a threshold $\theta$ is the discriminability of the object at $\theta$. (d) Scene images are represented by semantic scene descriptors (bottom), and clustering these descriptors exploit the semantic topics in fine-grained scene classes (top).
}
\label{fig:discriminative_objects}
\end{figure}

To this end, we make the following contributions.
We first propose a semantic scene descriptor that jointly captures the subtle differences between fine-grained scenes, while being robust to the different object configurations across domains. We compute the occurrence statistics of objects in scenes, capturing the informativeness of each detected object for each scene. We then transform such occurrences into scene probabilities.
This is complemented by a new measure of the discriminability of an object category, which is used to derive a discriminant dimensionality reduction procedure for object-based semantic representations. Second, we argue that scene images belong to multiple hidden semantic topics that can be automatically discovered by clustering our semantic descriptors. By learning a separate classifier for each discovered domain, the learnt classifiers are more discriminant. 
An overview of the proposed approach is shown in Figure \ref{fig:discriminative_objects}.

The third contribution is the introduction of the \textit{SnapStore} dataset, which addresses fine-grained scene classification with an emphasis
on robustness across imaging domains. It covers 18 visually-similar store categories, with training images downloaded from Google image search and test images collected with smartphones. To the best of our knowledge, SnapStore is the first dataset with these properties. It will be made publicly available from the author web-pages.

Finally, we compare the performance of the proposed method to state-of-the-art scene recognition and domain generalization methods. These show the effectiveness of the proposed scene transfer approach.

\section{Related work}
Recent approaches have been proposed to target domain generalization for vision tasks. They can be roughly grouped into classifier based \cite{undo_bias,lre_svm} approaches and feature-based \cite{dica,mtae} approaches. In \cite{undo_bias}, a support vector machine approach is proposed that learns a set of dataset-specific models and a visual-world model  that is common to all datasets. An exemplar-SVM approach is proposed in \cite{lre_svm} that exploits the structure of positive samples in the source domain. In feature-based approaches, the goal is to learn invariant features that generalize across domains. In \cite{dica}, a kernel-based method is proposed that learns a shared subspace. A feature-learning approach is proposed in \cite{mtae} that extends denoising autoeconders with naturally-occurring variability in object appearance. While the previous approaches yield good results in object recognition, their performance was not investigated for scene transfer. Also, to the best of our knowledge, there is no prior work that exploits a semantic approach to domain generalization.

Many approaches have been proposed for scene classification. A popular approach is to represent a scene in terms of its semantics \cite{semantic_spaces,semantic_manifold}, using a pre-defined vocabulary of visual concepts and a bank of detectors for those concepts \cite{object_bank,decaf,logFV,astounding_baseline,mop_cnn}.
A second class of approaches relies on the automatic discovery of  mid-level patches in scene images \cite{blocks_that_shout,discrim_patches,mode_seeking,d_parts}. While all these methods have been shown able to classify scenes, there are no previous studies of their performance for fine-grained classification.
Our method is most related to object-based approaches that are more suitable for fine-grained scenes than holistic representation methods, such as the scene gist \cite{gist}. Our proposed method is more invariant than previous attempts,
such as objectBank \cite{object_bank} and the semantic FV \cite{logFV}. These
methods provide an encoding based on raw (CNN-based)
detection scores, which vary widely across domains. In contrast, we quantize the detection scores into scene probabilities for each object. Such probabilities are adaptive to the varying detection scores through considering a range of thresholds. The process of quantization
imparts invariance to the CNN-based semantics, thus
improves the generalization ability.
We compare with both representations in Section \ref{sec:experiments}.

A Convolutional Neural Network \cite{alexnet,placesCNN},
is another example of a classifier that has the ability to discover
``semantic" entities in higher levels of its feature hierarchy~\cite{understanding_cnn,objects_emerge}.
The scene CNN of~\cite{placesCNN} was shown to detect objects that are discriminative for
the scene classes~\cite{objects_emerge}.
Our proposed method investigates scene transfer using a network trained on objects only, namely imageNET \cite{image_net}. This
is achieved without the need to train a network on millions of scene images, which is the goal of transfer. We compare the performance of
the two in Section \ref{sec:experiments}.

\section{SnapStore dataset}
\label{sec:dataset}
In order to study the performance of different methods for domain generalization for fine-grained scene recognition, we have assembled the SnapStore dataset.
This covers $18$ fine-grained \textit{store} categories, shown in Figure \ref{fig:dataset}.
Stores are a challenging scene classification domain for several reasons. First, many store categories have similar gist, i.e. similar global visual appearance and spatial layout.
For example, grocery stores, drug stores, and office supply stores all tend to contain long rows of shelves organized in a symmetric manner, with similar
floor and ceiling types. Second, store categories (e.g., clothing) that deviate from this norm, tend to exhibit a wide variation in visual appearance.
This implies that image models applicable to store classification must be detailed enough to differentiate among different classes of very similar visual appearance and
invariant enough to accommodate the wide variability of some store classes.

SnapStore contains $6132$ training images, gathered with Google image search. The number of training images per category varies from $127$ to $892$, with an average of $341$.
Training images were scaled to a maximum of $600$ pixels per axis. Testing images were taken in local stores, using smartphones. This results in images that are very different from those in the training set, which tend to be more stylized. The test set consists of $502$  images with ground truth annotations for store class, store location type
(shopping mall, street mall, industrial area), GPS coordinates, and store name.  Images have a fixed size of $960 \times 720$ pixels. Test images differ from training images in
geographical location, lighting conditions, zoom levels, and blurriness. This makes SnapStore a good dataset in which to test the robustness of scene classification to wide domain variations.

While datasets such as Places \cite{placesCNN} or SUN \cite{sun} contain some store categories, the proposed dataset is better suited for domain generalization of fine-grained scenes; first, SnapStore contains store classes that are more confusing, e.g., Drug store, DIY store, Office supplies store, and Multimedia store. Also, large datasets favor the use of machine learning methods that use data from the target domain to adapt to it.
In contrast, the images of SnapStore are explicitly chosen to stress robustness. This is
the reason why the test set includes images shot with cellphones,
while the training set does not. Overall, SnapStore
is tailored for the evaluation of representations and enables
the study of their robustness at a deeper level than Places
or SUN. We compare the three datasets in Section \ref{sec:experiments}.



\section{Discriminative objects in scenes}
There is a wide array of scenes that can benefit from object recognition, even if object cues are not sufficient for high recognition accuracy.
For example, we expect to see flowers in a flower shop, shoes and shoe boxes in a shoe shop, and chairs and tables in a furniture shop.
Nevertheless, it remains challenging to learn models that capture the discriminative power of objects for scene classification. First, objects can have different degrees of importance
for different scene types (e.g., chairs are expected in furniture stores, but also appear in shoe stores). Rather than simply accounting for the presence of an object in a scene,
there is a need to model how informative the object is of that scene. Second, object detection scores can vary widely across images, especially when these are from
different domains. In our experience, fixing the detection threshold to a value with good training performance frequently harms recognition accuracy on test images where the
object appears in different poses, different lighting, or occluded.

\subsection{Object detection and recognition}

An object recognizer $\rho: {\cal X} \rightarrow {\cal O}$ is a mapping from some feature space ${\cal X}$
to a set of object class labels ${\cal O}$, usually implemented as
$
  o = \arg \max_k f_k(x),
  \label{eq:recog}
$
where $f_k(x)$ is a confidence score for the assignment of a feature
vector $x \in {\cal X}$ to the $k^{th}$ label in ${\cal O}$.
An object detector is a special case, where ${\cal O} = \{-1,1\}$ and
$f_1(x) = -f_{-1}(x).$ In this case, $f_1(x)$ is simply denoted as
$f(x)$ and the decision rule of (\ref{eq:recog}) reduces to $ o = sgn[f(x)]$.

The function $f(x) = (f_1(x), \ldots, f_O(x)),$ where $O$ is the number of
object classes is usually denoted as the predictor of the recognizer
or detector. Component $f_k(x)$ is a {\it confidence score\/} for the
assignment of the object to the $k^{th}$ class. This is usually the
probability $P(o|x)$ or an invertible transformation of it.

Given an object recognizer, or a set of object detectors, it is
possible to detect the presence of object $o$ in an image $x$
at {\it confidence level\/} $\theta$ by thresholding the prediction
$f_o(x)$ according to
\begin{equation}
  \delta(x | o; \theta) = h[f_o(x) - \theta]
  \label{eq:delta}
\end{equation}
where $h(x) = 1, x \geq 0$ and $h(x) = 0$ otherwise.
Thus,  $\delta(x|o; \theta)$ is an indicator for the
assignment of image $x$ to object class $o$ at confidence level $\theta$.

\subsection{Learning an object occurrence model}
\label{sec:occ_models}
Our Object Occurrence Model (OOM) answers the following question on a threshold bandwidth of $[\theta_{min}; \theta_{max}]$ with a resolution of $\Delta \theta$: \textit{``how many images from each category contain the object at least once above a threshold $\theta$?''.}
We do not fix the threshold of object detection $\theta$ at a unique value as this threshold would be different across domains.
Formally, given a set ${\cal I}_c$ of images from a scene class $c$, the maximum
likelihood estimate of the probability of occurrence of object $o$ on class
$c$, at confidence level $\theta$, is
\begin{equation}
  p(o| c; \theta) = \frac{1}{|{\cal I}_{c}|}\sum_{x_i \in I_c}{\delta(x_i | o;
    \theta)}.
  \label{eq:probs}
\end{equation}
We refer to these probabilities, for a set of scene classes ${\cal C}$, as the object occurrence
model (OOM) of ${\cal C}$ at threshold $\theta$. This model summarizes the likelihood of
appearance of all objects in all scene classes, at this level of detection confidence.

\subsection{Discriminant object selection}
\label{sec:discriminativity}

\begin{figure*}[t]
\begin{center}
\includegraphics[width=\linewidth]{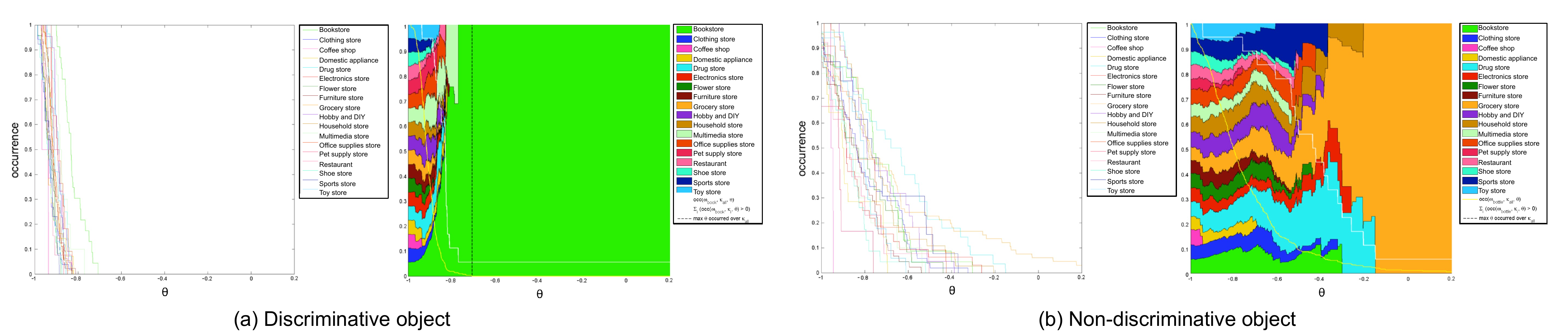}
\end{center}
\caption{An example of (a) a discriminative object (book) and (b) a
  non-discriminative object (bottle).
  In each case, the left plot is identical to the plot of
  Figure \ref{fig:discriminative_objects}c. The discriminative
  object (\textit{book}) occurs frequently in few categories at a given confidence level.
  However, for the same confidence level, the \textit{bottle} object,
  occurs in many categories.
  The plot on the right of (a) and (b) shows the occurrence normalized in
  1-norm for each $\theta$. The region above the
  maximal $\theta$ for any occurrence is interpreted as 1 for
  the category with the highest probability.}
\label{fig:discriminative_objects_example}
\end{figure*}

Natural scenes contain many objects, whose discriminative power varies
greatly. For example, the ``wall'' and ``floor'' objects are much
less discriminant than the objects ``pot,'' ``price tag,'' or
``flower'' for the recognition of ``flower shop'' images. To first order,
an object is discriminant for a particular scene class if it appears
frequently in that class and is uncommon in all others.
In general, an object can be discriminant for more than one class.
For example, the ``flower'' object is discriminant for the ``flower shop'' and
``garden'' classes.

We propose a procedure for discriminant object selection, based
on the OOM of the previous section. This relies on
a measure of the {\it discriminant power\/} $\phi_\theta(o)$ of object $o$ with
respect to a set of scene classes ${\cal C}$ at confidence level $\theta$.
The computation of  $\phi_\theta(o)$ is performed in two steps. First,
given object $o$, the classes $c \in {\cal C}$ are ranked according
to the posterior probabilities of (\ref{eq:post}).
Let $\gamma(c)$ be the ranking function, i.e. $\gamma(c) = 1$ for the
class of largest probability and $\gamma(c) = |{\cal C}|$ for the
class of lowest probability. The class of rank $r$ is then
$\gamma^{-1}(r)$. The second step computes the discriminant
power of object $o$ as
\begin{equation}
  \phi_\theta(o) = \max_{r \in \{1, \ldots, |{\cal C}|-1\}}
  p(\gamma^{-1}(r) | o; \theta) -
  p(\gamma^{-1}(r+1) | o; \theta).
\end{equation}

The procedure is illustrated in Figure~\ref{fig:discriminative_objects}c,
where each curve shows the probability $p(c|o; \theta)$ of class $c$
as a function of the confidence level. At confidence level $\theta$,
the red, green, yellow, and blue classes have rank $1$ to $4$ respectively.
In this example, the largest difference between probabilities
occurs between the green and yellow classes, capturing the fact that
the object $o$ is informative of the red and green classes but not of the
yellow and blues ones.

Figure \ref{fig:discriminative_objects_example} shows examples of
a discriminative and a non-discriminative object in the SnapStore dataset.
The discriminative object, \textit{book}, occurs in very few scene classes
(mainly bookstore) with high confidence level. On the other hand,
the non-discriminant  \textit{bottle} object appears in several classes
(grocery store, drug store, and household store) with the
same confidence level.

\section{Semantic latent scene topics}
In this section, we describe our approach of representing a scene image as scene probabilities, followed by discovering hidden semantic topics in scene classes.

\subsection{Semantic scene descriptor}
\label{sec:semantic_descriptors}
In this work, we propose to represent an image $x$ by a descriptor
based on the ${\cal O} \times {\cal C}$ matrix $M$ of posterior
probabilities $p(c|o)$ of classes given objects detected in the image.
Object detectors or recognizers produce multiple object detections in $x$, which are usually obtained by applying the recognizer or detector to image patches.
Object detectors are usually implemented in a
1-vs-rest manner and return the score of a binary decision.
We refer to these as hard detections. On the other hand, object
recognizers return a score vector, which summarizes the probabilities of
presence of each object in the patch. We refer to these as soft
detections.
Different types of descriptors are suitable
for soft vs. hard detections. In this work, we consider both,
proposing two descriptors that are conceptually identical but tuned
to the traits of the different detection approaches.

From the OOM, it is possible to derive the posterior probability of a scene class $c$
given the observation of object $o$ in an image $x$, at the confidence level $\theta$,
by simple application of Bayes rule
\begin{equation}
  p(c| o; \theta) =  \frac{p(o|c; \theta) p(c)}
  {\sum_ i p(o|i; \theta) p(i)},
  \label{eq:post}
\end{equation}
where $p(o|c; \theta)$  are the probabilities of occurrence of~(\ref{eq:probs}) and
$p(c)$ is a prior scene class probability.
The range of thresholds $[\theta_{\min}, \theta_{\max}]$ over which
$\theta$ is defined is denoted the {\it threshold bandwidth\/} of the model.
\subsubsection{Hard detections}
Given the image
$x$, we apply to it the $i^{th}$ object detector, producing
a set of $n_i$ bounding boxes, corresponding to image patches
${\cal X}_i = \{z^{(i)}_1, \ldots, z^{(i)}_{n_i}\}$, and a set
of associated detection
scores ${\cal S}_i = \{s^{(i)}_1, \ldots, s^{(i)}_{n_i}\}$.
To estimate the posterior probabilities $p(c|o_i)$, we adopt
a Bayesian averaging procedure, assuming that these scores are
samples from a probability distribution $p(\theta)$ over
confidence scores. This leads to
$p(c|o_i) = \sum_k p(c|o_i, \theta = s^{(i)}_k) p(\theta = s^{(i)}_k).$
Assuming a uniform prior over scores, we then use $p(\theta = s^{(i)}_k) = 1/n_i$ to obtain
\begin{equation}
  p(c|o_i) = \frac{1}{n_i}\sum_k p(c|o_i, \theta = s^{(i)}_k).
\end{equation}
In summary, the vector of posterior probabilities is estimated by averaging
the OOM posteriors of~(\ref{eq:post}), at the confidence levels associated
with the object detections in $x$. This procedure is repeated for all
objects, filling one row of $M$ at a time. The rows associated with
undetected objects are set to zero.

The proposed  semantic descriptor is obtained by stacking $M$ into a
vector and performing \textit{discriminant} dimensionality reduction. We start by finding an object
subset ${\cal R} \subset {\cal O}$ which is discriminant for scene
classification. This reduces dimensionality
from $|{\cal O}| \times |{\cal C}|$ to $|{\cal R}| \times |{\cal C}|$ as
discussed in Section~\ref{sec:discriminativity}. 
This procedure is repeated using a spatial pyramid
structure of three levels ($1\times1$, $2\times2$, and  $3\times1$), which are
finally concatenated into a $21K$ dimensional feature vector.

\subsubsection{Soft detections}

A set of $n$ patches
${\cal X} = \{z_1, \ldots, z_{n}\}$ are sampled from the
image and fed to an object recognizer, e.g. a CNN.
This produces a set ${\cal S} = \{s_1, \ldots, s_n\}$
of vectors $s_k$ of confidence scores. The vector $s_k$
includes the scores for the presence of all $|{\cal O}|$ objects in patch
$z_k$. Using the OOM posteriors of~(\ref{eq:post}), each $s_k$ can be converted into a matrix $M^k$ of class
probabilities given scores. Namely the matrix whose $i^{th}$ row is given by
$M_i^K = p(c | o_i, s_{k,i})$, which
is the vector of class probabilities given the detection
of object $o_i$ at confidence $s_{k,i}$.

The image $x$ is then represented as a
bag of descriptors ${\cal X} = \{M^1, M^2, \hdots M^n\}$
generated from its patches. This is mapped into the soft-VLAD~\cite{vlad,mop_cnn} representation using the following steps.
First, the dimensionality of the matrices $M^k$ is reduced by selecting
the most discriminant objects $\cal{R} \subset \cal{O}$, as discussed in
Section~\ref{sec:discriminativity}. Second, each matrix is stacked into
a ${\cal R} \times {\cal C}$ vector, and dimensionality
reduced to $500$ dimensions, using PCA. The descriptors are then encoded with the soft-kmeans assignment
weighted first order residuals, as suggested in~\cite{mop_cnn}.


\subsection{Semantic clustering}
\label{sec:semantic_clustering}
When learning knowledge from web data or multiple datasets, it is usually assumed that training images may come from several hidden topics \cite{wsdg,lre_svm} that may correspond to different viewing angles, or imaging conditions. While previous works rely on image features like DeCaF fc6 \cite{decaf} to discover latent topics in \textit{object} datasets, we instead propose to discover \textit{semantic} topics that provide a higher level of abstraction, which generalizes better than lower-level features especially for \textit{scene} datasets.
Each of the hidden topics can contain an arbitrary number of images from an arbitrary number of scene classes. For example, furniture store images can be semantically divided into different groups, as shown in Figure \ref{fig:discriminative_objects}, including 1) images of dining furniture that are semantically related to some images in `Coffee Shop' and `Restaurant' classes, 2) images of seating furniture, like sofas and ottomans, that are related to waiting areas in `Shoe shop' class, and 3) images of bedroom furniture that are more unique to furniture stores. By exploiting such underlying semantic structure of fine-grained classes, we achieve better discriminability by learning a separate multi-class classifier for each latent topic. Furthermore, improved generalization ability is achieved through integrating the decisions from all the learnt classifiers at test time \cite{merge_classifiers}. This is especially useful when the test image does not fall uniquely into one of the topics as is usually common in cross-domain settings.
We note that our goal is to project the training images into a semantic space that can yield informative groups when clustered using any clustering method, not necessarily k-means.

In practice, we first partition the training data into $D$ semantic latent topics using k-means clustering over our semantic descriptors (Section \ref{sec:semantic_descriptors}) from all training images. Note that we do not assume any underlying distribution in the data and we do not utilize scene labels in discovering the latent topics. We then learn a classifier $f_{c,d}(\textbf{x})$ for each class $c$ in each latent topic $d$ using only the training samples in that domain. 
The classifier models of each latent topic are learnt using 1-vs-rest SVM with linear kernel,
using the \texttt{JSGD} library \cite{jsgd}. The regularization
parameter and learning rate were determined by 5-fold cross
validation.
At test time, we predict the scene class of an image \textbf{$x$} as the class with the highest decision value after average pooling the classifier decisions from all topics, by using $y = arg \max_c \sum_{d=1}^{D}{f_{c,d}(\textbf{x})}$. We also experimented with max pooling over classifier decisions, which yielded inferior results.

\section{Experiments}
\label{sec:experiments}
A number of experiments were designed to evaluate the performance of the proposed method. All datasets are weakly labeled -
scene class labels, no object bounding boxes - and
we report average classification accuracy over scene classes.
In all experiments, hard object detections were obtained with
the RCNN of~\cite{rcnn} and soft detections with the CNN
of~\cite{alexnet}. We empirically fix $k=5$ for $k$-means clustering (Sec. \ref{sec:semantic_clustering}), however the results are insensitive to the exact value of $k$.

\subsection{Analysis of the object occurrence model (OOM)}
In this experiment, we used the new SnapStore dataset, which addresses fine-grained classification, and
MIT67 \cite{indoor_scenes}, which addresses coarse-grained indoor scenes.
The latter includes $67$ indoor scene categories. We used 
the train/test split proposed by the authors, using $80$ training 
and $20$ test images per class. 
\begin{figure}[t]
\begin{center}
\begin{subfigure}[b]{0.49\textwidth}
          \includegraphics[width=\linewidth]{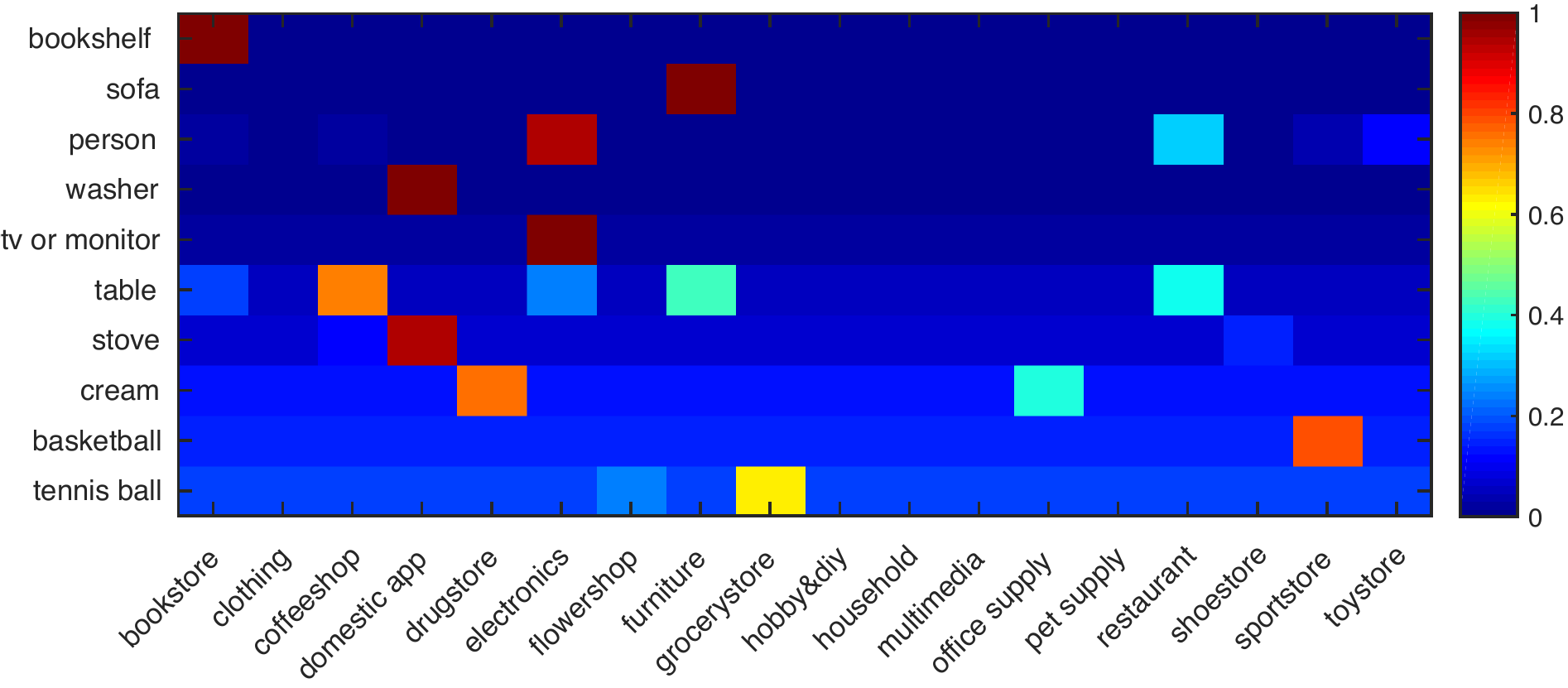}
        \caption{ }
        \label{fig:occ_snapstore_best}
\end{subfigure}
\begin{subfigure}[b]{0.49\textwidth}
  \includegraphics[width=\linewidth]{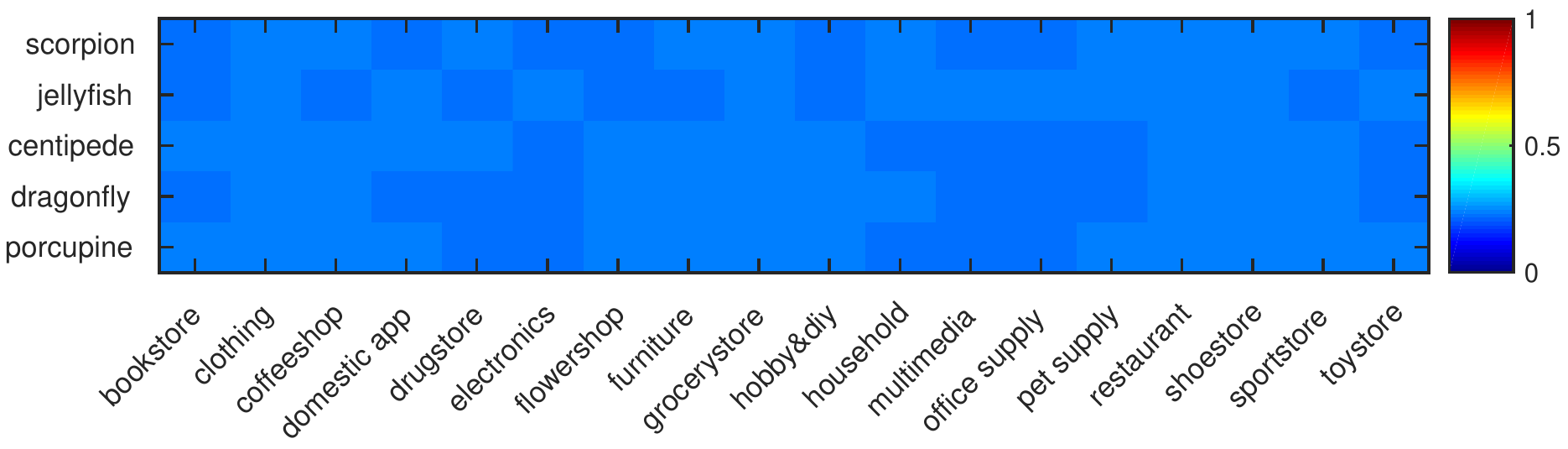}
  \caption{ }
 \label{fig:occ_snapstore_worst}
 \end{subfigure}
\end{center}
 \caption{
   Scene likelihoods for all scene classes for (a) the  top 10 discriminative objects and (b) the least discriminative objects using RCNN-200 on SnapStore}
\label{fig:occ_snapstore}
\end{figure}

\begin{figure}[t]
\begin{center}
\begin{subfigure}[b]{0.47\textwidth}
          \includegraphics[width=\linewidth]{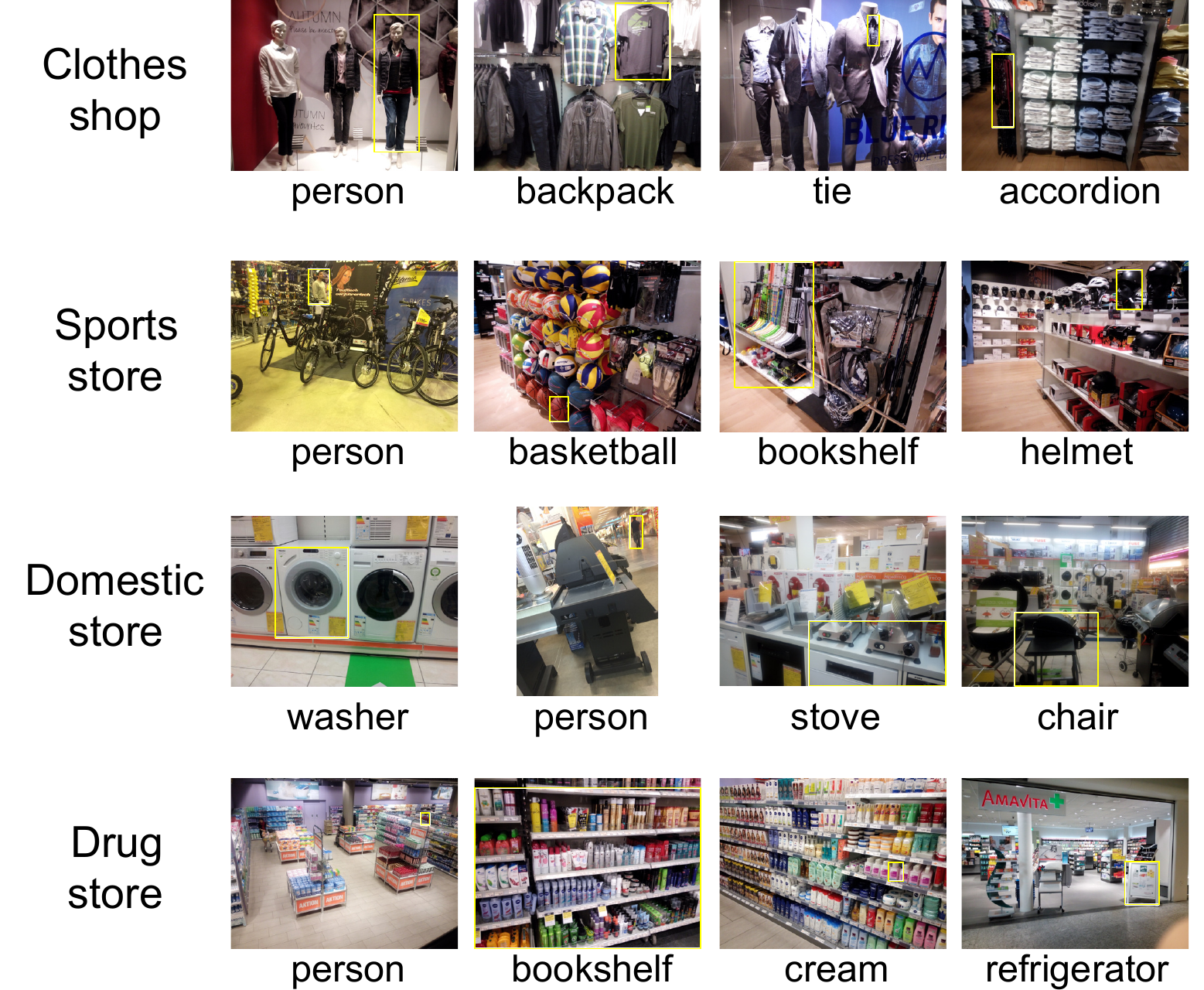}
        \caption{ }
\end{subfigure}
\begin{subfigure}[b]{0.47\textwidth}
  \includegraphics[width=\linewidth]{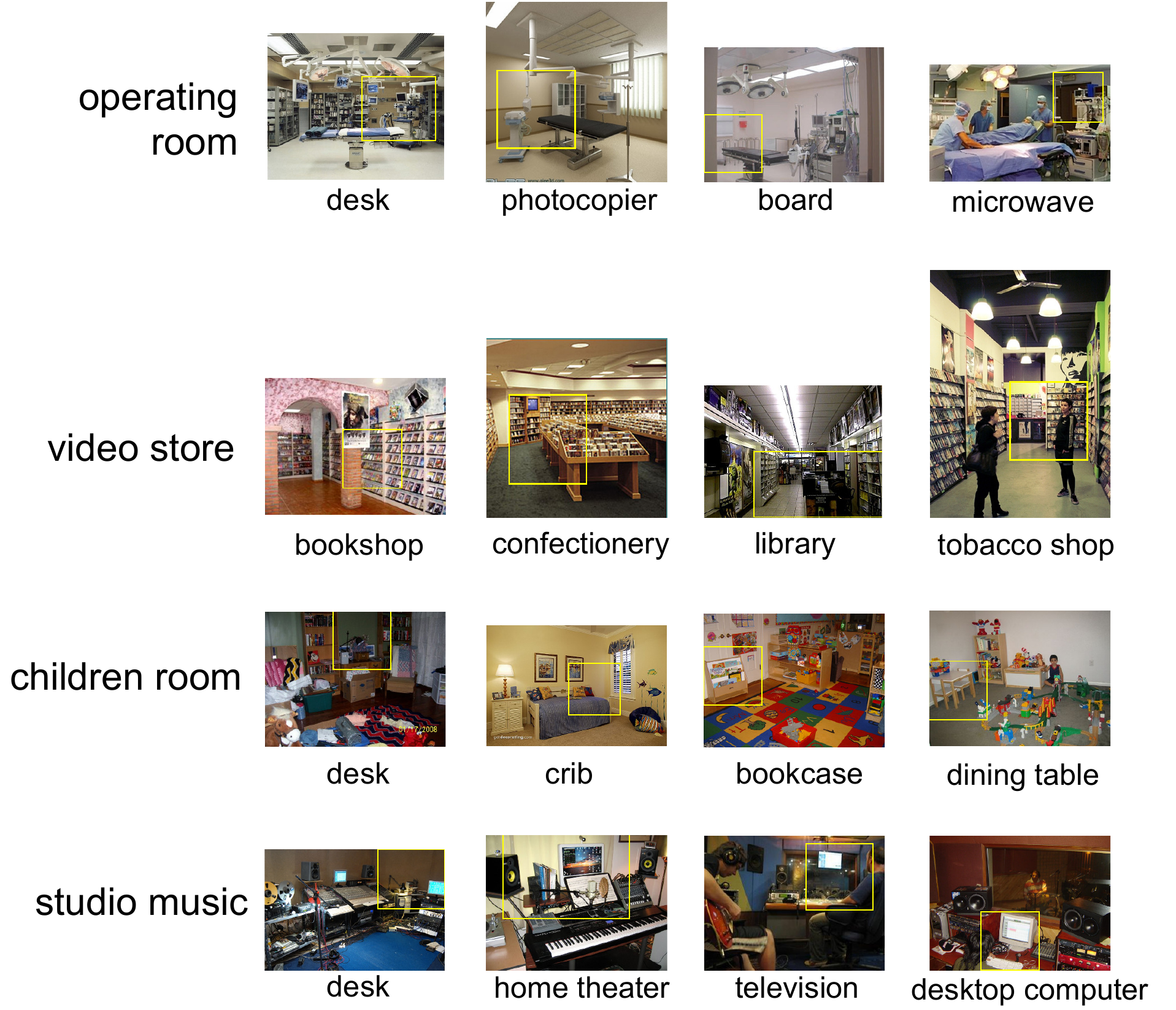}
  \caption{ }
 \end{subfigure}
\end{center}
   \caption{Scene categories of higher recognition rate for (a) hard detections
     on SnapStore, and (b) soft detections on MIT67. 
     }
\label{fig:qualitative_results}
\end{figure}

\begin{table}[t]
\caption{Classification accuracy as a function of the number of
  discriminant objects for SnapStore and MIT67}
\begin{center}
\label{table:object_selection_results}
\begin{tabular}{c|c|c|c}
\hline
Dataset & OOM \small{[CNN-1000]}  & OOM \small{[CNN-500]} & OOM \small{[CNN-300]} \\
\hline
SnapStore & 43.1 & 44.6 & \textbf{45.4} \\
\hline
MIT67 & 68.0 & \textbf{68.2} & 66.4 \\
\hline
\end{tabular}
\end{center}
\end{table}

Figure \ref{fig:occ_snapstore_best} shows the matrix of posterior class
probabilities learned by the OOM, for hard detections
on SnapStore. A similar plot is shown in the supplement for detections 
on MIT67. The figure shows a heatmap of the probabilities
$p(c|o_i; \theta)$ of~(\ref{eq:post}) at the confidence 
level $\theta = 0.9$. Note that the OOM captures the informative objects 
for each scene class, e.g., bookshelf is highly discriminant for the 
bookstore class. Furthermore, when an object is discriminant for multiple 
classes, the class probabilities reflect the relative importance of the 
object, e.g., table is discriminant for coffee shops, furniture stores, 
and restaurants but more important for the coffee shop class. While
nearly all coffee shop images contain tables, furniture store images 
sometimes depict beds, sofas or other objects, and some pictures of 
fast-food restaurant lack tables. 
Figure \ref{fig:occ_snapstore_worst} shows the same heatmap for the least 
discriminant objects. The scene probabilities are now identical for all
objects, which are hardly detected in any of the scenes. 

Figure \ref{fig:qualitative_results} shows the top four 
correctly-classified scene classes on SnapStore and MIT67. 
Scene classes are sorted from top to bottom by decreasing classification 
accuracy. For each scene, we show the most probable objects (most common 
object on the left) along with the bounding box of highest detection score. 
While there are noisy detections in each class, e.g. accordion in clothes 
shop, as a whole the detections are
quite informative of the scene class. Failure cases on SnapStore include 
multimedia store, office supply store, and toy store. 

We investigated the performance as a function of the number of selected discriminant objects (Section \ref{sec:discriminativity}).  
Table \ref{table:object_selection_results} summarizes the performance
of soft-detections (CNN) without semantic clustering, when using different numbers of objects. For both datasets, the selection of discriminant objects
is beneficial, although the gains are larger in SnapStore. 
Using a reduced object vocabulary also reduces the dimensionality of
the descriptors, leading to more efficient classification. 
For hard detections on SnapStore, we observed a similar improvement 
of performance for reduction from the $200$ object vocabulary of the RCNN 
to $140$ objects. On MIT67, the $200$ object vocabulary proved
inadequate to cover the diversity of objects in the 67 scene classes.
Given these results, we fixed the number of objects at $140$ for 
hard-detections (RCNN) and $300$  for soft detections (CNN) on SnapStore. 
On MIT67, we used $200$ and $500$ objects, respectively.

\subsection{Cross recognition performance on SnapStore dataset}

\begin{table} [t]
\caption{Comparison of  classification accuracies on SnapStore. 
  \small{*-Indicates results for a single scale of 
    $128\times128$ patches}}
\begin{center}
\begin{tabular}{c|c}
\hline
Method & Accuracy (\%)  \\
\hline
GIST \cite{gist} & 22.8 \\
DiscrimPatches \cite{discrim_patches} & 25.0 \\
ObjectBank \cite{object_bank} & 32.6 \\
\hline
ImageNET finetune &  38.6\\
ImagetNET fc7 + SVM (DeCaF) \cite{decaf} & 40.2 \\
Places finetune & 42.4 \\
Places fc7  & 44.2 \\
\hline
ObjectBank [CNN]* & 34.8\\
ObjectBank [RCNN] & 36.3\\
fc8-VLAD (semantic FV) \cite{logFV}* & 43.8 \\
\hline
DICA \cite{dica} & 24.2 \\
\hline
OOM \small{[CNN]}* \small{ (Ours)} & 45.4\\
OOM \small{[RCNN] (Ours)} & 45.7 \\
OOM-semanticClusters \small{[RCNN] (Ours)} & \textbf{47.9} \\
\hline
\end{tabular}
\end{center}
\label{table:snapstore_results}
\end{table}

We performed a comparison to state-of-the-art scene recognition and transfer methods on the \textbf{18} classes of SnapStore in Table \ref{table:snapstore_results}. We additionally compare with ObjectBank \cite{object_bank} when using RCNN and CNN detections as our method in exactly the same settings, to perform a fair comparison with it.
We cannot compare with Undo-Bias \cite{undo_bias} as it requires the source domains to be explicitly associated with multiple datasets. We compare with their method in Section \ref{sec:experiments_multiple_datasets}.

OOM with RCNN outperformed all other  methods, including a finetuned Places CNN.
Semantic clustering further improves the recognition by $\approx 2\%$.
Note that Places fc7 is trained on scenes, while we use a network trained on objects only, which shows successful scene transfer.
Places fine-tune surprisingly yielded worse performance than Places fc7. This is because Places fine-tune overfits to training views, performing better on images from the training domain, but worse on the new domain.
Our method improves over ObjectBank by $\approx$ 9\%, when using CNN detectors and recognizers. This is attributed to our invariant representation that does not rely on raw detection scores, which are different
across domains. The small dimensionality of the DICA descriptor limits its discriminative ability. 

\subsection{Cross recognition performance on multiple datasets}
\label{sec:experiments_multiple_datasets}
Here, we evaluate the effectiveness of the proposed algorithm when using multiple fine-grained scene datasets. We also study the bias in each dataset, showing the benefits of using SnapStore to test the robustness of recognition methods. 
\\ 
\textbf{Datasets.} We used images from the \textbf{9} fine-grained store scene classes that are common among SnapStore, SUN \cite{sun}, and Places \cite{placesCNN} datasets. 
Effectively, we have $4$ datasets, each divided into training and validation sets. The class names and detailed training-test configuration are provided in the supplement.
\\ 
\textbf{Baselines.} We compared two variants of our method, namely OOM on RCNN (\textbf{OOM}) and OOM on RCNN + semantic clustering (\textbf{OOM-SC}), with 6 baselines: \textbf{DeCaF}, DeCaF + $k$-means clustering (\textbf{DeCaF-C}), Undo-Bias\cite{undo_bias} (\textbf{U-B}), \textbf{DICA}\cite{dica}, ObjectBank on RCNN (\textbf{OB}), and ObjectBank on RCNN + our proposed semantic clustering (\textbf{OB-SC}). For DeCaF-C, we set $k$ = 2, which yielded the best results for this method. Note that we cannot compare with Places CNN in this experiment as it was trained using millions of images from Places dataset, thus violating the conditions of domain generalization on \textit{unseen} datasets.
\\ 
\textbf{Results.} To show the dataset bias and evaluate the ground truth performance, we first measured the cross-recognition performance of a linear SVM on DeCaF fc7 features when using the training set of one dataset and the test set of another dataset. We summarize the results in Table \ref{table:multiple_datasets}. Results show a significant bias in datasets gathered from the web (SnapWeb, SUN, Places). This is shown by the significant drop in performance by $>12\%$ when using SnapPhone dataset, which is gathered in real settings using a smartphone, as the testing set. In contrast, the cross-recognition performance when using SUN and Places datasets as train/test sets is much better, with only $3\%$ drop in performance when compared to ground truth (same-domain) recognition. This emphasizes the benefits of using the proposed SnapStore dataset in evaluating scene transfer methods.

\begin{table}[t]
\caption{Ground truth and cross-recognition accuracy (\%) of DeCaF+SVM baseline on multiple fine-grained scene datasets}
\begin{center}
\begin{tabular}{c|c|c|c||c}
\hline
Training/Test & SUN & SnapWeb & Places & SnapPhone  \\
\hline
SUN & \textbf{68.7} & 57.1 & 65.7 & 56.5 \\
SnapWeb & 62.7 & \textbf{71.9} & 60.9 & 58.2 \\
Places & 64.2 & 59.2 & \textbf{67.6} & 53.8 \\
\hline
\end{tabular}
\end{center}
\label{table:multiple_datasets}
\end{table}

\begin{table}[t]
\caption{Cross-recognition accuracy (\%) on SnapStore training set (SnW), SnapStore test set (SnP), SUN, and Places (Pla) datasets}
\begin{center}
\begin{tabular}{c|c||c|c|c|c|c|c|c|c}
\hline
Train & Test &  DeCaF &  DeCaF-C & U-B & DICA & OB & OB-SC & \textbf{OOM} & \textbf{OOM-SC}  \\
\hline
SnW & SnP & 58.2 & 56.3 & N/A & 42.1 & 30.0 & 37.4 & 61.1 & \textbf{62.0} \\
SUN & SnP & 56.5 & 53.9 & N/A & 45.5 & 39.2 & 35.9  & 54.4 & \textbf{56.9} \\
Pla & SnP & 53.8 & 49.1 & N/A & 37.7 & 27.6 & 28.3 & \textbf{54.8} & 54.6 \\
\hline
SnW,SnP & Pla,SUN &  59.1 & 59.9 & 52.3 & 49.2 & 22.7 & 25.7 & 57.3 & \textbf{60.6} \\
SnW,SUN & SnP,Pla & 60.6 & 58.5 & 50.3 & 52.2 & 37.4 & 37.7 & 61.0 & \textbf{63.2} \\
SUN,Pla,SnW & SnP &  59.7 & 57.2 & 47.8 & 53.5 & 36.3 & 39.1 & 61.6 & \textbf{62.5} \\
SUN,SnP,SnW & Pla &  \textbf{63.8} & 62.2 & 33.8 & 50.8 & 27.4 & 30.2 & 59.8 & 63.3\\
\hline\hline
\multicolumn{2}{c||}{Average} & 58.8 & 56.7 &46.0& 47.2 & 32.9 & 33.4 & 58.5 & \textbf{60.4} \\
\hline
\end{tabular}
\end{center}
\label{table:cross_dataset}
\end{table}

We then evaluated the cross-recognition performance of the proposed method and the baselines, as summarized in Table \ref{table:cross_dataset}. Our method outperforms other methods on five out of seven cross-domain scenarios and on average.
The improvement of the proposed approach over DeCaF is more significant in the experiment in Section 6.2. This is due to the similarity of images in SUN, Places, and SnW, all collected on the web, which benefits the DeCaF baseline. When testing on SnP even OOM beats DeCaF on 3 of 4 cases with an average of 58\% vs. 57\%.
Clustering DeCaF features (DeCaF-C) yielded worse performance than the DeCaF baseline. This is because DeCaF features are spatial maps that discriminate between parts of objects or at most individual objects.  Thus, clustering them produces clusters of visually similar object parts, limiting invariance against varying object poses and shapes across domains.
Recent work \cite{decaf_clusters} made similar observations about DeCaF clusters for object datasets.
One interesting observation is the inferior performance of domain generalization methods.
While such methods yielded impressive performance on object datasets, they are unsuitable for fine-grained scenes;
Undo-Bias associates a source domain to each source dataset, which does not capture the semantic topics across the scene classes, while the small dimensionality of the DICA descriptor limits its discriminability.

\subsection{Scene recognition on coarse-grained and same domain dataset}

\begin{table}[t]
\caption{Comparison of  classification accuracies on MIT67. 
  \small{*-Indicates results for a single scale of 
    $128\times128$ patches}.}
\begin{center}
\begin{tabular}{c|c}
\hline
Method & Accuracy (\%)  \\
\hline
IFV \cite{blocks_that_shout} & 60.7  \\
MLrep \cite{mode_seeking} & 64.0 \\
\hline
DeCaF \cite{decaf} &58.4  \\
ImageNET finetune & 63.9 \\
OverFeat + SVM \cite{astounding_baseline} & 69  \\
fc6 + SC \cite{fc6SC} & 68.2  \\
fc7-VLAD \cite{mop_cnn} [4 scales/1 scale*] & 68.8 / 65.1 \\
\hline
ObjectBank [RCNN / CNN*] & 41.5 / 48.5 \\
fc8-FV \cite{logFV} [ 4 scales/1 scale*] & 72.8 / 68.5 \\
\hline
OOM \small{[RCNN]} \small{(Ours)} & 49.4 \\
OOM \small{[CNN]*} \small{(Ours)} & 68.2\\
OOM-semClusters \small{(Ours)} & 68.6 \\
\hline
\end{tabular}
\end{center}
\label{table:results_mit67}
\end{table}

Finally, we compared the performance to state-of-the-art scene recognition methods on the coarse-grained MIT67 dataset in Table \ref{table:results_mit67}. Soft detections achieved
the best performance. The performance of hard-detections was rather weak,
due to the limited vocabulary of the RCNN.
We achieve comparable performance to state-of-the-art scene recognition algorithms, which shows that the effectiveness of the proposed method is more pronounced in cross-domain settings. 

\section{Conclusion}
In this work, we proposed a new approach for domain generalization for fine-grained scene recognition. To achieve robustness against varying object configurations in scenes across domains, we quantize object occurrences into conditional scene probabilities. We then exploit the underlying semantic structure of our representation to discover hidden semantic topics. We learn a disriminant classifier for each domain that captures the subtle differences between fine-grained scenes. SnapStore, a new dataset of fine-grained scenes in cross-dataset settings was introduced. Extensive experiments have shown the effectiveness of the proposed approach and the benefits of SnapStore for fine-grained scene transfer.

\bibliographystyle{splncs}
\bibliography{egbib}

\begin{thebibliography}{10}

\bibitem{indoor_scenes}
Quattoni, A., Torralba, A.:
\newblock Recognizing indoor scenes.
\newblock In: CVPR. (2009)

\bibitem{stuff}
Adelson, E.H.:
\newblock On seeing stuff: the perception of materials by humans and machines.
\newblock Proc. SPIE \textbf{4299} (2001)  1--12

\bibitem{ucsd_birds}
Welinder, P., Branson, S., Mita, T., Wah, C., Schroff, F., Belongie, S.,
  Perona, P.:
\newblock Caltech-\text{UCSD} birds 200.
\newblock \text{Technical Report CNS-TR-201}, Caltech (2010)

\bibitem{oxford_flowers}
Nilsback, M.E., Zisserman, A.:
\newblock Automated flower classification over a large number of classes.
\newblock In: ICVGIP. (2008)

\bibitem{dataset_bias}
Torralba, A., Efros, A.:
\newblock Unbiased look at dataset bias.
\newblock In: CVPR. (2011)

\bibitem{bias_perronnin_2010}
Perronnin, F., Senchez, J., Liu, Y.:
\newblock Large-scale image categorization with explicit data embedding.
\newblock In: CVPR. (2010)

\bibitem{domain_adaptation_survey}
Patel, V.M., Gopalan, R., Li, R., Chellappa, R.:
\newblock Visual domain adaptation: A survey of recent advances.
\newblock In: IEEE Signal Processing Magazine. (2014)

\bibitem{domain_adaptation_1}
Bruzzone, L., Marconcini, M.:
\newblock Domain adaptation problems: A {DASVM} classification technique and a
  circular validation strategy.
\newblock PAMI \textbf{32} (2010)  770--787

\bibitem{domain_adaptation_2}
Duan, L., Tsang, I.W., Xu, D.:
\newblock Domain transfer multiple kernel learning.
\newblock PAMI \textbf{34} (2012)  465--479

\bibitem{domain_adaptation_3}
Baktashmotlagh, M., Harandi, M., Lovell, M.S.B.:
\newblock Unsupervised domain adaptation by domain invariant projection.
\newblock In: ICCV. (2013)

\bibitem{domain_adaptation_4}
Fernando, B., Habrard, A., Sebban, M., Tuytelaars, T.:
\newblock Unsupervised visual domain adaptation using subspace alignment.
\newblock In: ICCV. (2013)

\bibitem{dica}
Muandet, K., Balduzzi, D., Scholkopf, B.:
\newblock Domain generalization via invariant feature representation.
\newblock In: ICML. (2013)

\bibitem{undo_bias}
Khosla, A., Zhou, T., Malisiewicz, T., Efros, A.A., Torralba, A.:
\newblock Undoing the damage of dataset bias.
\newblock In: ECCV. (2012)

\bibitem{lre_svm}
Xu, Z., Li, W., Niu, L., Xu, D.:
\newblock Exploiting low-rank structure from latent domains for domain
  generalization.
\newblock In: ECCV. (2014)

\bibitem{mtae}
Ghifary, M., Kleijn, W.B., Zhang, M., Balduzzi, D.:
\newblock Domain generalization for object recognition with multi-task
  autoencoders.
\newblock In: ICCV. (2015)

\bibitem{wsdg}
Niu, L., Li, W., Xu, D.:
\newblock Visual recognition by learning from web data: A weakly supervised
  domain generalization approach.
\newblock In: CVPR. (2015)

\bibitem{semantic_spaces}
Rasiwasia, N., Vasconcelos, N.:
\newblock Scene classification with low-dimensional semantic spaces and weak
  supervision.
\newblock In: CVPR. (2008)

\bibitem{semantic_manifold}
Kwitt, R., Vasconcelos, N., Rasiwasia, N.:
\newblock Scene recognition on the semantic manifold.
\newblock In: ECCV. (2012)

\bibitem{object_bank}
Li, L.J., Su, H., Xing, E.P., {Fei-Fei}, L.:
\newblock Object bank: A high-level image representation for scene
  classification and semantic feature sparsification.
\newblock In: NIPS. (2010)

\bibitem{decaf}
Donahue, J., Jia, Y., Vinyals, O., Hoffman, J., Zhang, N., Tzeng, E., Darrell,
  T.:
\newblock {DeCaF}: A deep convolutional activation feature for generic visual
  recognition.
\newblock In: ICML. (2014)

\bibitem{logFV}
Dixit, M., Chen, S., Gao, D., Rasiwasia, N., Vasconcelos, N.:
\newblock Scene classification with semantic fisher vectors.
\newblock In: CVPR. (2015)

\bibitem{astounding_baseline}
Razavian, A.S., Azizpour, H., Sullivan, J., Carlsson, S.:
\newblock {CNN} features off-the-shelf: An astounding baseline for recognition.
\newblock In: CVPR Workshops. (2014)

\bibitem{mop_cnn}
Gong, Y., Wang, L., Guo, R., Lazebnik, S.:
\newblock Multi-scale orderless pooling of deep convolutional activation
  features.
\newblock In: ECCV. (2014)

\bibitem{blocks_that_shout}
Juneja, M., Vedaldi, A., Jawahar, C.V., Zisserman, A.:
\newblock Blocks that shout: Distinctive parts for scene classification.
\newblock In: CVPR. (2013)

\bibitem{discrim_patches}
Singh, S., Gupta, A., Efros, A.A.:
\newblock Unsupervised discovery of mid-level discriminative patches.
\newblock In: ECCV. (2012)

\bibitem{mode_seeking}
Doersch, C., Gupta, A., Efros, A.:
\newblock Mid-level visual element discovery as discriminative mode seeking.
\newblock In: NIPS. (2013)

\bibitem{d_parts}
Sun, J., Ponce, J.:
\newblock Learning discriminative part detectors for image classification and
  cosegmentation.
\newblock In: ICCV. (2013)

\bibitem{gist}
Oliva, A., Torralba, A.:
\newblock Modeling the shape of the scene: A holistic representation of the
  spatial envelope.
\newblock IJCV \textbf{42}(3) (2001)  145--175

\bibitem{alexnet}
Krizhevsky, A., Sutskever, I., Hinton, G.E.:
\newblock {ImageNet} classification with deep convolutional neural networks.
\newblock In: NIPS. (2012)

\bibitem{placesCNN}
Zhou, B., Lapedriza, A., Xiao, J., Torralba, A., , Oliva, A.:
\newblock Learning deep features for scene recognition using places database.
\newblock In: NIPS. (2014)

\bibitem{understanding_cnn}
Zeiler, M.D., Fergus, R.:
\newblock Visualizing and understanding convolutional networks.
\newblock In: ECCV. (2014)

\bibitem{objects_emerge}
Zhou, B., Khosla, A., Lapedriza, {\`{A}}., Oliva, A., Torralba, A.:
\newblock Object detectors emerge in deep scene cnns.
\newblock CoRR \textbf{abs/1412.6856} (2014)

\bibitem{image_net}
Deng, J., Dong, W., Socher, R., Li, L.J., Li, K., {Fei-Fei}, L.:
\newblock Imagenet: A large-scale hierarchical image database.
\newblock In: CVPR. (2009)

\bibitem{sun}
Xiao, J., Hays, J., Ehinger, K., Oliva, A., Torralba, A.:
\newblock {SUN} database: Large-scale scene recognition from abbey to zoo.
\newblock In: CVPR. (2010)

\bibitem{vlad}
J\'egou, H., Douze, M., Schmid, C., P\'erez, P.:
\newblock Aggregating local descriptors into a compact image representation.
\newblock In: CVPR. (2010)

\bibitem{merge_classifiers}
Kittler, J., Hatef, M., Duin, R.P.W., Matas, J.:
\newblock On combining classifiers.
\newblock PAMI \textbf{20}(3) (1998)  226--239

\bibitem{jsgd}
Akata, Z., Perronnin, F., Harchaoui, Z., Schmid, C.:
\newblock Good practice in largescale learning for image classification.
\newblock PAMI \textbf{36}(3) (2013)  507--520

\bibitem{rcnn}
Girshick, R., Donahue, J., Darrell, T., Malik, J.:
\newblock Rich feature hierarchies for accurate object detection and semantic
  segmentation.
\newblock In: CVPR. (2014)

\bibitem{decaf_clusters}
Tzeng, E., Hoffman, J., Zhang, N., Saenko, K., Darrell, T.:
\newblock Deep domain confusion: Maximizing for domain invariance.
\newblock In: {CoRR}, abs/1412.3474. (2014)

\bibitem{fc6SC}
Liu, L., Shen, C., Wang, L., {van den Hengel}, A., Wang, C.:
\newblock Encoding high dimensional local features by sparse coding based
  {F}isher vectors.
\newblock In: NIPS. (2014)

\end{thebibliography}
\end{document}